\documentclass[10pt,twocolumn,letterpaper]{article}

\usepackage{iccv}
\usepackage{times}
\usepackage{epsfig}
\usepackage{graphicx}
\usepackage{amsmath}
\usepackage{amssymb}

\newcommand{\minisection}[1]{\vspace{1mm}\noindent{\bf #1}}
\usepackage{booktabs}
\usepackage[caption=false,font=footnotesize]{subfig}

\usepackage[pagebackref=true,breaklinks=true,letterpaper=true,colorlinks,bookmarks=false]{hyperref}

\iccvfinalcopy 

\def\iccvPaperID{0019}

\ificcvfinal\fi
\begin{document}

\title{FacePoseNet: Making a Case for Landmark-Free Face Alignment}

\author{Feng-Ju Chang\textsuperscript{1}, Anh Tuan Tran\textsuperscript{1}, Tal Hassner\textsuperscript{2,3}, Iacopo Masi\textsuperscript{1}, Ram Nevatia\textsuperscript{1}, Gerard Medioni\textsuperscript{1}\\
\textsuperscript{1}~Institute for Robotics and Intelligent Systems, USC, CA, USA\\
\textsuperscript{2}~Information Sciences Institute, USC, CA, USA\\
\textsuperscript{3}~The Open University of Israel, Israel\\
{\tt\small \{fengjuch,anhttran,iacopoma,nevatia,medioni\}@usc.edu, hassner@isi.edu}
}

\maketitle

\begin{abstract}
We show how a simple convolutional neural network (CNN) can be trained to accurately and robustly regress 6 degrees of freedom (6DoF) 3D head pose, directly from image intensities. We further explain how this FacePoseNet (FPN) can be used to align faces in 2D and 3D as an alternative to explicit facial landmark detection for these tasks. We claim that in many cases the standard means of measuring landmark detector accuracy can be misleading when comparing different face alignments. Instead, we compare our FPN with existing methods by evaluating how they affect face recognition accuracy on the IJB-A and IJB-B benchmarks: using the same recognition pipeline, but varying the face alignment method. Our results show that (a) better landmark detection accuracy measured on the 300W benchmark does not necessarily imply better face recognition accuracy. (b) Our FPN provides superior 2D and 3D face alignment on both benchmarks. Finally, (c), FPN aligns faces at a small fraction of the computational cost of comparably accurate landmark detectors. For many purposes, FPN is thus a far faster and far more accurate face alignment method than using facial landmark detectors.
\end{abstract}

\section{Introduction}
Facial landmark detection is rarely, if ever, an application in its own right. Instead, it is typically a means to an end: It is one component out of many in pipelines designed for other face understanding and processing tasks, often providing effective means for aligning face photos and making them easier to process. Most facial landmark detectors, however, are developed without measuring their impact on these applications but rather using standard facial landmark detection benchmarks such as the popular AFW~\cite{zhu2012face}, LFPW~\cite{belhumeur2013localizing}, HELEN~\cite{le2012interactive}, and IBUG~\cite{sagonas2013300}. These benchmarks contain face images with manually labeled {\em ground truth} landmarks. Better detection accuracy on these benchmarks equals better prediction of these manual positions. This raises an important question: {\em Does better approximation of such human labeled landmarks imply better face alignment and consequently better face understanding?}

\begin{figure}[t]
\centering
\includegraphics[width=.95\linewidth]{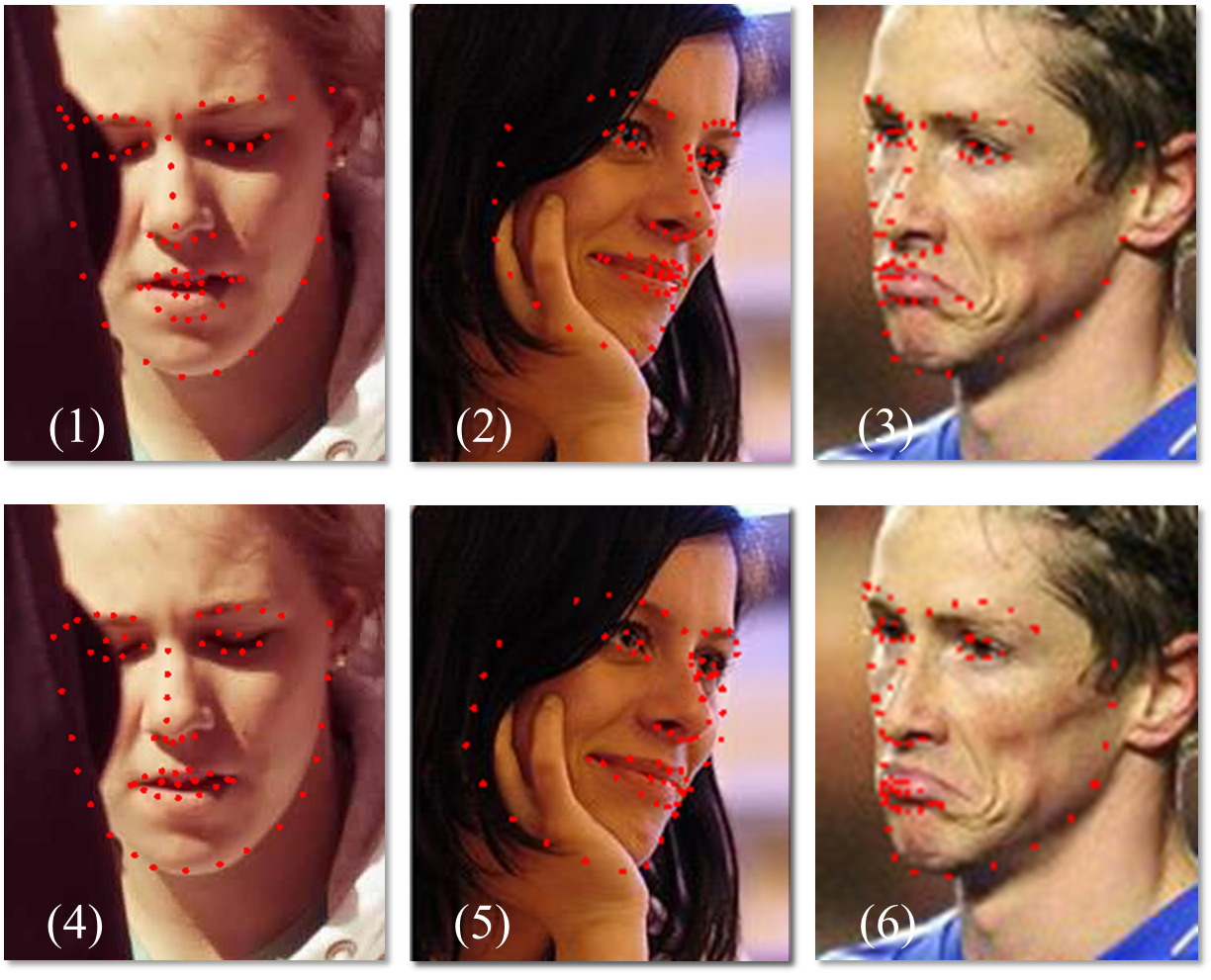}
\caption[foo]{
{\em The problem with manually labeled ground truth facial landmarks.} Images and annotations from the AFW~\cite{zhu2012face} (left two columns) and iBug~\cite{sagonas2013300} benchmarks. One of each pair shows  manually labeled ground truth landmarks; the other, a high-error prediction of our FPN, which does not account for facial expression or 3D shape. Which is which?\footnotemark[1] Clearly, detection accuracy, as measured by standard benchmarks, does not necessarily reflect the quality of the landmark detection.}\label{fig:gt_vs_fpn}
\end{figure}

\footnotetext[1]{\rotatebox{180}{Images one, three, and five are ground truth.}}

Why would higher accuracy on landmark detection benchmarks {\em not} imply better alignment? The many landmark detection benchmarks used by the community to measure detection accuracy typically offer 5, 49 or 68 landmarks painstakingly labeled on hundreds or thousands of unconstrained face images, reflecting wide viewpoint, resolution and noise variations. On low resolution images, however, even expert human operators can find it hard to accurately pinpoint landmark positions. More importantly, many landmark locations are not well defined even in high resolution (e.g., points along the jawline or behind occlusions). Thus, improved landmark detection accuracy may actually reflect better estimation of uncertain human labels rather than better face alignment (Fig.~\ref{fig:gt_vs_fpn}).

An additional concern relates to how landmarks are used for face alignment. Face alignment often implies using a global 2D or 3D transformation to {\em warp} faces to ideal, reference frames: Detected landmarks are matched with their corresponding landmarks in the reference coordinates and a 2D or 3D transformation is then computed by robust estimation methods. To our knowledge, the effects landmark detection noise, changing expressions or face shapes have on these estimated transformations were never fully explored. 

Responding to these concerns, we offer several contributions. (1) We propose comparing landmark detection methods by evaluating bottom line face recognition accuracy on faces aligned with these methods. (2) As an alternative to existing facial landmark detectors, we further present a robust and accurate, landmark-free method for face alignment: our deep FacePoseNet (FPN). We show it to excel at global, 3D face alignment even under the most challenging viewing conditions. Finally, (3), we test our FPN extensively and report that better landmark detection accuracy on the widely used 300W benchmark~\cite{sagonas2015300} does {\em not} imply better alignment and recognition on the highly challenging IJB-A~\cite{Klare_2015_CVPR} and IJB-B benchmarks~\cite{whitelam2017iarpa}. In particular, recognition results on images aligned with our FPN surpass those on images aligned with state-of-the-art detectors.  

Some applications require landmark estimation. Our FPN provides a more accurate and far faster face alignment technique in the many cases where global alignment, rather than specific landmark positions, is needed. To support our claims, we make our code publicly available~\footnote{\url{https://github.com/fengju514/Face-Pose-Net}}.

\section{Related work}\label{sec:related}
\minisection{Applications of facial landmark detectors.}
Facial landmark detection is big business, as reflected by the numerous citation to relevant papers, the many facial landmark detection benchmarks~\cite{belhumeur2013localizing,kostinger2011annotated,le2012interactive,sagonas2015300,zhu2012face}, and popular international events dedicated to this problem. With all this effort, a rigorous survey of the many applications of facial landmarks is outside the scope of this paper. In lieu of such a survey, and to get some idea of why this problem attracts so much attention, we offer the following cursory study.

\begin{figure}[t]
\centering
\includegraphics[width=\linewidth]{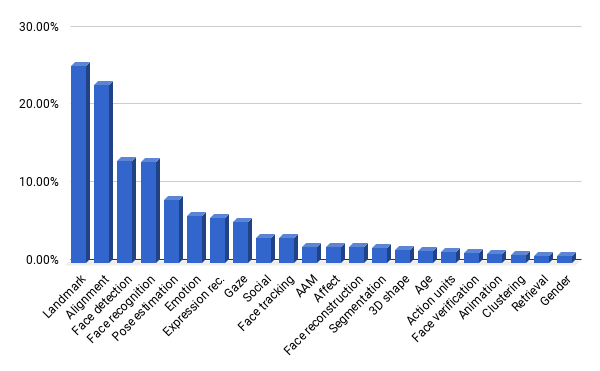}
\caption{
{\em Applications of facial landmarks.} Illustrating the frequency of various task and application names in paper titles citing two of the most popular landmark detectors~\cite{zhu2012face} and~\cite{xiong2013supervised}.}\label{fig:wordfreq}
\end{figure}

We consider two of the most widely cited face landmark detector papers of the last decade, the tree based approach~\cite{zhu2012face} and supervised descent method~\cite{xiong2013supervised}. At the time of writing, based on Google Scholar, the latter accumulated nearly a thousand citations and the former well over a thousand. We found 23 application names appearing frequently (more than ten times) in the titles of the papers that cite these two and counted the number of times these applications were mentioned. The relative frequencies of these applications are reported in Fig.~\ref{fig:wordfreq}.

Of course, this simple survey is by no means accurate: the same term is counted twice if the paper using it in its title cites both~\cite{zhu2012face} and~\cite{xiong2013supervised} and many paper titles do not clearly state the application they describe (e.g.,~\cite{hassner2013viewing} describes a method for face alignment in 3D but does not mention ``alignment'' in the title). Nevertheless, with around two thousand papers included in this survey, the result is quite clear: Alignment, face recognition and pose estimation -- also considered alignment -- are {\em overwhelmingly more popular} than any other application. This, of course, excluding other landmark detection papers. 

 \minisection{What does it mean to align a face?}
The term {\em alignment} almost always appears in the titles of papers which present facial landmark detection methods~\cite{asthana2014incremental,cao2014face,ren2014face} (and most others) implying that the two terms are used interchangeability. This reflects an interpretation of alignment as {forming correspondences between particular spatial locations in one face image and another}. A different interpretation of {\em alignment}, and the one used here, refers not only to establishing these correspondences but also to {\em warping} the two face images, thus making them easier to compare and match. Face warping with estimated 2D (in-plane) or 3D transformations is well known to have a profound impact on the performance of face recognition systems~\cite{hassner2015effective,huang2007unsupervised}. 

Although sometimes alignments involve non-parametric or part-based warps~\cite{hassner2013viewing,hassner2013single}, often, global 2D or 3D (parametric) transformation are all that is required for this purpose. Such aligned faces are then further processed in systems for face recognition~\cite{chen2014cross,ding2016multi,zhu2015high}, emotion recognition~\cite{huang2014improved,levi2015emotion}, age and gender estimation~\cite{eidinger2013age,gurpinar2016kernel,LH:CVPRw15:age}, and more. In fact, it was recently claimed that a global alignment is both more robust and far faster to warp than non-parametric transformations~\cite{masi2017rapid,masi16dowe}. This paper focuses on such global transformations, showing how they can be estimated quickly and accurately using a deep neural network.

\minisection{Deep pose estimation.} This work describes a deep network trained to estimate the 6DoF of 3D faces viewed in single images. Deep learning is increasingly used for similar purposes, though typically focusing on general object classes~\cite{bansal2016marr,poirson2016fast,su2015render}. Some recently addressed faces in particular, though their methods are designed to estimate 2D landmarks along with 3D face shapes~\cite{jourabloo2016large,7961750,zhu2015}. Unlike our proposed pose estimation, they regress poses by using iterative methods which involve computationally costly face rendering. We regress 6DoF directly from image intensities without such rendering steps.

In all these cases, absence of training data was cited as a major obstacle for training effective models. In response, some turned to larger 3D object data sets~\cite{xiang2014beyond,xiang2016objectnet3d} or using synthetically generated examples~\cite{richardson2016learning}. We propose a far simpler alternative and show it to result in robust and accurate face alignment.

\section{A critique of facial landmark detection}\label{sec:critique}
Before using an existing state-of-the-art facial landmark detector in a face processing system, the following points should be considered.

\minisection{Landmark detection accuracy measures.} Facial landmark detection accuracy is typically measured by considering the distances between estimated landmarks and ground truth (reference) landmarks, normalized by the reference inter-ocular distance of the face~\cite{dantone2012real}:
\begin{equation}
{e}(\mathbf{L},\hat{\mathbf{L}})= \frac{1}{m\|{\hat{\mathbf{p}}}_{l}-{\hat{\mathbf{p}}}_{r}\|_2} \sum_{i=1}^{m} \| \mathbf{p_i} - \hat{\mathbf{p_i}} \|_2,\label{eq:error}
\end{equation}
\noindent Here, $\mathbf{L} = \{\mathbf{p_i}\}$ is the set of $m$ 2D facial landmark coordinates, $\hat{\mathbf{L}} = \{\hat{\mathbf{p_i}}\} $ their ground truth locations, and ${\hat{\mathbf{p}}}_{l}, {\hat{\mathbf{p}}}_{r}$ the reference left and right eye outer corner positions. These errors are then translated to a number of standard quantities, including the mean error rate (MER), the percentage of landmarks detected under certain error thresholds (e.g., below 5\% or 10\% error rates) or the area under the accumulative error curve (AUC). 

There are two key problems with this method of evaluating landmark errors. First, the ground truth compared against is manually specified, often by mechanical turk workers. These manual annotations can be noisy, they are ill-defined when images are low resolution, the landmarks are occluded (in case of large out-of-plane head rotations, facial hair and other obstructions), or located in featureless facial regions (e.g., along the jawline). Accurate facial landmark detection, as measured on these benchmarks, thus implies better matching human labels but not necessarily better detection. These problems are demonstrated in Fig.~\ref{fig:gt_vs_fpn}.

\begin{figure*}[t]
\centering
\includegraphics[width=\linewidth]{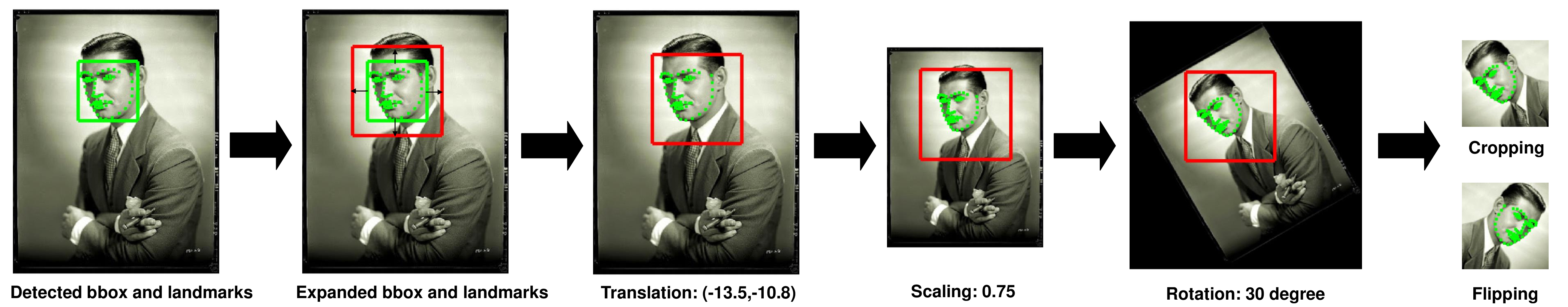}
\caption{
{\em Augmenting appearances of images from the VGG face dataset~\cite{Parkhi15}.} After detecting the face bounding box and landmarks we augment its appearance by applying a number of simple planar transformations, including translation, scaling, rotation, and flipping. The same transformations are applied to the landmarks, thereby producing example  landmarks for images which may be too challenging for existing landmark detectors to process.\vspace{-3mm}}\label{fig:perturb}
\end{figure*}

A second potential problem lies in the error measure itself: Normalizing detection errors by inter-ocular distances biases against images of faces appearing at non-frontal views. When faces are near profile, perspective projection of the 3D face onto the image plane shrinks the distances between the eyes thereby naturally inflating the errors computed for such images.

\minisection{Landmark detection speed.} Some facial landmark detection methods emphasize impressive speeds~\cite{king2009dlib,ren2014face}. Measured on standard landmark detection benchmarks, however, these methods do not necessarily claim state-of-the-art accuracy, falling behind more sophisticated, yet far slower detectors~\cite{zadeh2016deep}. Moreover, aside from~\cite{zhu2015}, no existing landmark detector is designed to take advantage of GPU hardware, a standard feature in commodity computer systems and most, including~\cite{zhu2015}, apply iterative optimizations which may be hard to convert to parallel processing.

\minisection{Effects of facial expression and shape on alignment.} It was recently shown that 3D alignment and warping of faces to frontal viewpoints (i.e. {\em frontalization}) is effective regardless of the precise 3D face shape used for this purpose~\cite{hassner2015effective}. Facial expressions and 3D shapes in particular, appear to have little impact on the warped result as evident by the improved face recognition accuracy reported by that method. Moreover, it was recently demonstrated that by using such a generic 3D face shape, rendering faces from new viewpoints can be accelerated to the same speed as simple 2D image warping~\cite{masi2017rapid}.

Interestingly, they and many others used facial landmark detectors to compute parametric transformations -- projection matrix~\cite{hassner2015effective} or 2D affine or similarity transforms~\cite{eidinger2013age,huang2007unsupervised} -- by applying robust estimators to corresponding detected facial landmarks~\cite{hassner2013viewing,lepetit2009epnp}. Variations in landmark locations due to expressions and face shapes essentially {\em contribute noise} to this estimation process. The effects these variations have on the quality of the alignment were, as far as we know, never truly studied.

\section{Deep, direct head pose regression}\label{sec:methodmain}
Rather than align faces using landmark detection, we refer to alignment as a global, 6DoF 3D face pose, and propose to infer it directly from image intensities, using a simple deep network architecture. We next describe the network and the novel method used to train it.

\subsection{Head pose representation}\label{sec:training}
We define face alignment as the 3D head pose $\mathbf{h}$, expressed using 6DoF: three for rotations, $\mathbf{r}=(r_x,r_y,r_z)^T$, and three for translations, $\mathbf{t}=(t_x,t_y,t_z)^T$:
\begin{equation}
\mathbf{h} = (r_x, r_y, r_z, t_x, t_y, t_z)^T
\end{equation}
where $(r_x, r_y, r_z)$ are represented as Euler angles (pitch, yaw, and roll). Given $m$ 2D facial landmark coordinates on an input image, $\mathbf{p}_{2\times m}$, and their corresponding, reference 3D coordinates, $\mathbf{P}_{3\times m}$ -- selected on a fixed, generic 3D face model -- we can obtain a 3D to 2D projection of the 3D landmarks onto the 2D image by solving the following equation for the standard pinhole model:
\begin{equation}
[\mathbf{p},\mathbf{1}]^T = \mathbf{A}[\mathbf{R},\mathbf{t}][\mathbf{P},\mathbf{1}]^T,
\label{eqn:pinhole_model}
\end{equation}
where $\mathbf{A}$ and $\mathbf{R}$ are the camera matrix and rotation matrix respectively and $\mathbf{1}$ is a constant vector of 1. We then extract a rotation vector $\mathbf{r}=(r_x,r_y,r_z)^T$ from $\mathbf{R}$ using the Rodrigues rotation formula:
$$
\mathbf{R} = \cos\mathbf{\theta}\mathbf{I} + (1-\cos\mathbf{\theta})\mathbf{r}\mathbf{r}^T + \sin\mathbf{\theta}
\begin{pmatrix} 
0 & -r_z & r_y \\ 
r_z & 0 & -r_x \\ 
-r_y & r_x & 0  
\end{pmatrix},
$$
where we define $\mathbf{\theta} = ||\mathbf{r}||_2$.

\minisection{Obtaining enough training examples.}
Although our network architecture is not very deep compared to deep networks used today  for other tasks, training it still requires large quantities of labeled training data. We found the numbers of facial landmark annotated faces in standard data sets to be too small for this purpose. A key problem is therefore obtaining a large enough training set.

We produce our training set by synthesizing 6D, ground truth pose labels by running an existing facial landmark detector~\cite{baltruvsaitis2016openface} on a large image set: the 2.6 million images in the VGG face dataset~\cite{Parkhi15}. The detected landmarks were then used to compute the 6DoF labels for the images in this set. A potential danger in using an existing method to produce our training labels, is that our CNN will not improve beyond the accuracy of its training labels. As we show in Sec.~\ref{sec:results}, this is not necessarily the case.

To further improve the robustness of our CNN, we apply a number of face augmentation techniques to the images in the VGG face set, substantially enriching the appearance variations it provides. Fig.~\ref{fig:perturb} illustrates this augmentation process. Specifically, following face detection~\cite{yang2016multi} and landmark detection~\cite{baltruvsaitis2016openface}, we transform detected bounding boxes and their detected facial landmarks using a number of simple in-plane transformations. The parameters for these transformations are selected randomly from fixed distributions (Table.~\ref{tab:perturb_dist}). The transformed faces are then used for training, along with their horizontally mirrored versions, to provide yaw rotation invariance. Ground truth labels are, of course, computed using the transformed landmarks. 

\begin{figure*}[t]
\centering
\includegraphics[width=\linewidth]{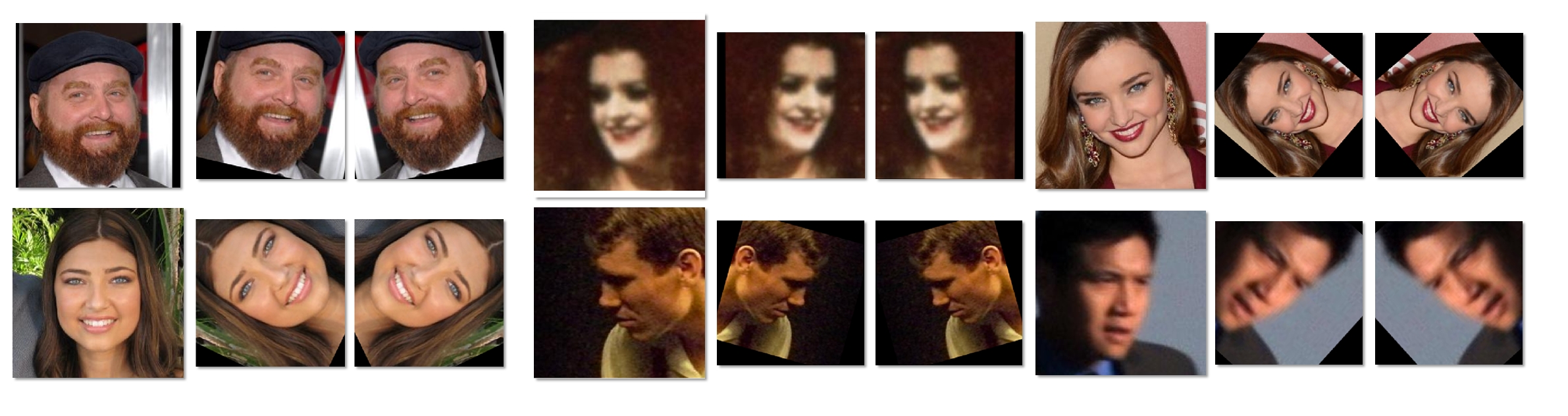}
\caption{
{\em Example augmented training images.} Example images from the VGG face data set~\cite{Parkhi15} following data augmentation. Each triplet shows the original detected bounding box (left) and its augmented versions (mirrored across the vertical axis). Both flipped versions were used for training FPN. Note that in some cases, detecting landmarks would be highly challenging on the augmented face, due to severe rotations and scalings not normally handled by existing methods. Our FPN is trained with the original landmark positions, transformed to the augmented image coordinate frame.\vspace{-3mm}}\label{fig:augmented}
\end{figure*}

Some example augmented faces are provided in Fig.~\ref{fig:augmented}. Note that augmented images would often be {\em too challenging for existing landmark detectors}, due to extreme rotations or scaling. This, of course, {\em does not affect the accuracy of the ground truth labels} which were obtained from the original images. It does, however, force our CNN to learn to estimate poses even on such challenging images.

\begin{table}[t]
\caption{{\em Summary of augmentation transformation parameters used to train our FPN.} Where $\mathcal{U}(a,b)$ samples from a uniform distribution ranging from $a$ to $b$ and $\mathcal{N}(\mu,\sigma^2)$ samples from a  normal distribution with mean $\mu$ and variance $\sigma^2$. $width$ and $height$ are the face detection bounding box dimensions.}
\label{tab:perturb_dist}
\begin{center}
\begin{tabular}{lc}
\toprule
\textbf{Transformation} & \textbf{Range}\\ \hline
Horizontal translation & $\mathcal{U}(-0.1,0.1) \times width$ \\
Vertical translation & $\mathcal{U}(-0.1,0.1) \times height$ \\
Scaling & $\mathcal{U}(0.75,1.25)$ \\
Rotation (degrees)& $30\times \mathcal{N}(0,1)$ \\
\bottomrule
\end{tabular}
\end{center}
\end{table}

\minisection{FPN training.}
For our FPN we use an AlexNet architecture~\cite{AlexNet} with its initialized weights provided by~\cite{masi2016pose}. The only difference is that here the output regresses 6D floating point values rather than predicts one-hot encoded, multi class labels. Note that during training each dimension of the head pose labels is normalized by the corresponding mean and standard deviation of the training set, compensating for the large value differences among dimensions. The same normalization parameters are used at test time. 

\minisection{2D and 3D face alignment with FPN.}
Given a test image, it is processed by applying the same face detector~\cite{yang2016multi}, cropping the face and scaling it to the dimension of the network's input layer. The 6D network output is then converted to a projection matrix. Specifically, the projection matrix is produced by the camera matrix $\mathbf{A}$, rotation matrix $\mathbf{R}$, and the translation vector $\mathbf{t}$ in Eq.~(\ref{eqn:pinhole_model}). With this projection matrix we can render new views of the face, aligning it across 3D views as was recently proposed by others~\cite{masi16dowe,masi2017rapid}.

For 2D alignment, we compute the 2D similarity transform to warp the 2D projected landmarks to pre-defined landmark locations. With frontal images (absolute yaw angle $ \le 30^\circ$), we use the eye centers, the nose tip, and the mouth corners for alignment. With profile images (absolute yaw angle $ > 30^\circ$), however, only the visible eye center and the nose tip are used.

\section{Results}\label{sec:results}
We provide comparisons of our FPN with the following widely used, state-of-the-art, facial landmark detection methods: Dlib~\cite{king2009dlib}, CLNF~\cite{baltrusaitis2013constrained}, OpenFace~\cite{baltruvsaitis2016openface}, DCLM~\cite{zadeh2016deep}, RCPR~\cite{burgos2013robust}, and 3DDFA~\cite{zhu2015} evaluating them for their effects on face recognition vs. their landmark detection accuracy. 

\subsection{Effect of alignment on recognition}\label{sec:IJBAB}
Sec.~\ref{sec:critique} discusses the various potential problems of comparing face alignment methods by measuring their landmark detection accuracy. As an alternative, we propose comparing methods for face alignment and landmark detection by evaluating their effect on the bottom line accuracy of a face processing pipeline. Since face recognition is arguably one of the most popular applications for face alignment, we use recognition accuracy as a performance measure. To our knowledge, this is the first time alignment methods are compared based on their effect on recognition accuracy.

\begin{table}[t]
\caption{{\em Verification and identification on IJB-A and IJB-B,} comparing landmark detection based face alignment methods. Three baseline IJB-A results are also provided as reference at the top of the table. $^{*}$ Numbers estimated from the ROC and CMC in~\cite{whitelam2017iarpa}.}
\label{tab:IJBAB}
\setlength{\tabcolsep}{1.5em}
\begin{center}
\resizebox{\linewidth}{!}{
\begin{tabular}{l@{~~}c@{~~}c@{~~}c@{~~}c@{~~}c@{~~}c@{~~}c@{~~}c}
\toprule
\textbf{Method $\downarrow$} & \multicolumn{3}{c}{\textbf{TAR@FAR}}  &  \multicolumn{4}{c}{\textbf{Identification Rate (\%)}}\\ 
Eval. $\rightarrow$  & .01\%~ & 0.1\%~ & 1.0\%~ & Rank-1 & Rank-5 & Rank-10 & Rank-20   \\ \hline
\multicolumn{8}{c}{\textbf{IJB-A}~\cite{Klare_2015_CVPR}}\\ \hline
Crosswhite et al.~\cite{crosswhite2017template} & --  & --  & 93.9  & 92.8  & --  & 98.6  & -- \\ %\hline
Ranjan et al.~\cite{ranjan2017l2} & 90.9  & 94.3  & 97.0  & 97.3  & --  & 98.8  & -- \\
Masi et al.~\cite{masi2017rapid}$$  & 56.4  & 75.0  & 88.8  & 92.5  & 96.6  & 97.4  & 98.0 \\ \hline
RCPR~\cite{burgos2013robust}  & 64.9  & 75.4  & 83.5  & 86.6  & 90.9  & 92.2  & 93.7\\ 
Dlib~\cite{king2009dlib}  & 70.5  & 80.4  & 86.8  & 89.2  & 91.9  & 93.0  & 94.2 \\ 
CLNF~\cite{baltrusaitis2013constrained}  & 68.9 & 75.1 & 82.9 & 86.3 & 90.5 & 91.9 & 93.3 \\
OpenFace~\cite{baltruvsaitis2016openface}  & 58.7 & 68.9 & 80.6 & 84.3 & 89.8 & 91.4 & 93.2 \\ 
DCLM~\cite{zadeh2016deep}  & 64.5 & 73.8 & 83.7 & 86.3 & 90.7 & 92.2 & 93.7 \\
3DDFA~\cite{zhu2015}  & {74.8} &	{82.8} &	{89.0} &	90.3 &	{92.8} &	{93.5} & {94.4} \\ \hline
Our FPN  & \textbf{77.5}  & \textbf{85.2}  & \textbf{90.1}  & \textbf{91.4}  & \textbf{93.0}  & \textbf{93.8}  & \textbf{94.8} \\  \bottomrule
\multicolumn{8}{c}{\textbf{IJB-B}~\cite{whitelam2017iarpa}}\\ \toprule
GOTs~\cite{whitelam2017iarpa}$^{*}$ & 16.0 & 33.0 & 60.0 & 42.0 & 57.0 & 62.0 & 68.0 \\ 
VGG face~\cite{whitelam2017iarpa}$^{*}$ & 55.0 & 72.0 & 86.0 & 78.0 &  86.0 & 89.0 & 92.0 \\ \hline
RCPR~\cite{burgos2013robust} & 71.2 & 83.8 & 93.3 & 83.6 & 90.9 & 93.2 & 95.0\\ 
Dlib~\cite{king2009dlib} & 78.1 & 88.2 & 94.8 & 88.0 & 93.2 & 94.9 & 96.3 \\ 
CLNF~\cite{baltrusaitis2013constrained} & 74.1 & 	85.2 & 	93.4 & 	84.5 & 	90.9 & 	93.0 & 	94.8 \\
OpenFace~\cite{baltruvsaitis2016openface} & 54.8 & 	71.6 & 	87.0 & 	74.3 & 	84.1 & 	87.8 & 	90.9 \\ 
DCLM~\cite{zadeh2016deep} & 67.6 & 81.0 & 92.0 & 81.8 & 89.7 & 92.0 & 94.1 \\
3DDFA~\cite{zhu2015} & 78.5 & 	89.1 & 	{95.6} & 	89.0 & 	94.1 & 	95.5 & 	{96.9} \\  \hline
Our FPN & \textbf{83.2} & \textbf{91.6} & \textbf{96.5} & \textbf{91.1} & \textbf{95.3} & \textbf{96.5} & \textbf{97.5} \\ 
\bottomrule
\end{tabular}
}
\end{center}
\end{table}

\begin{figure*}[t]
\centering
\begin{tabular}{cc}
\includegraphics[width=.42\linewidth]{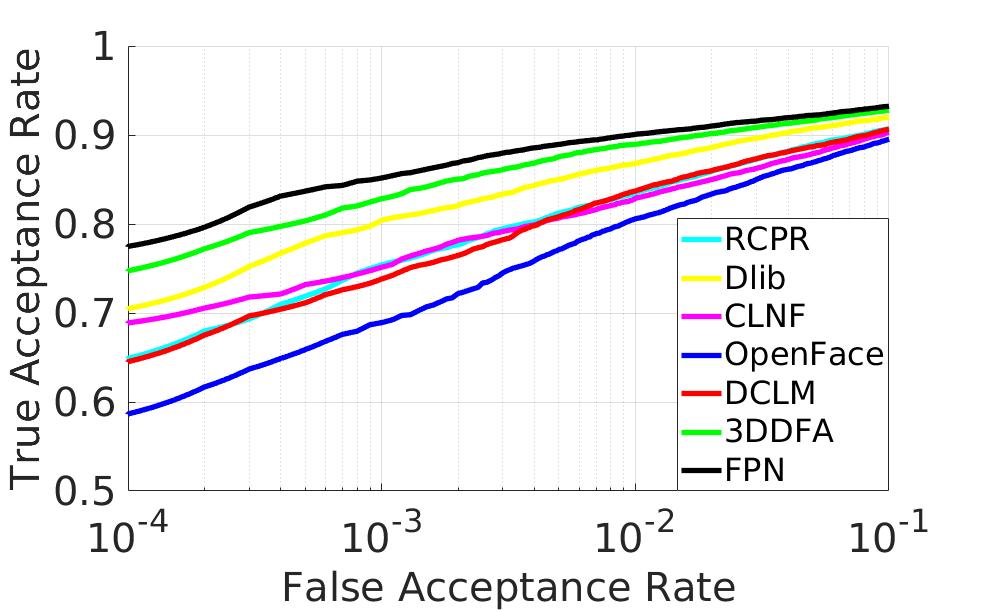}&
\includegraphics[width=.42\linewidth]{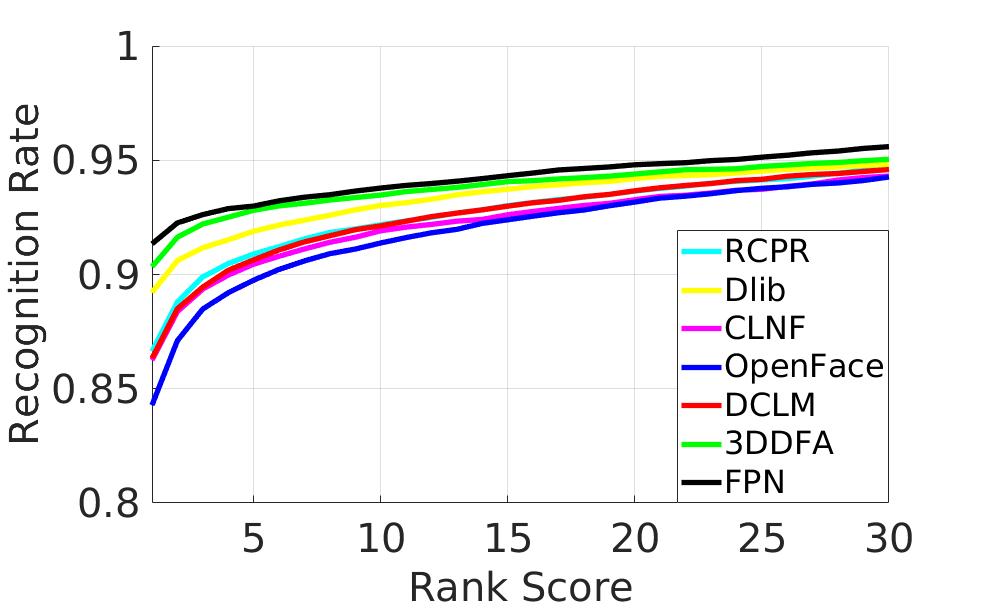}\\
(a) ROC IJB-A & (b) CMC IJB-A\\
\includegraphics[width=.42\linewidth]{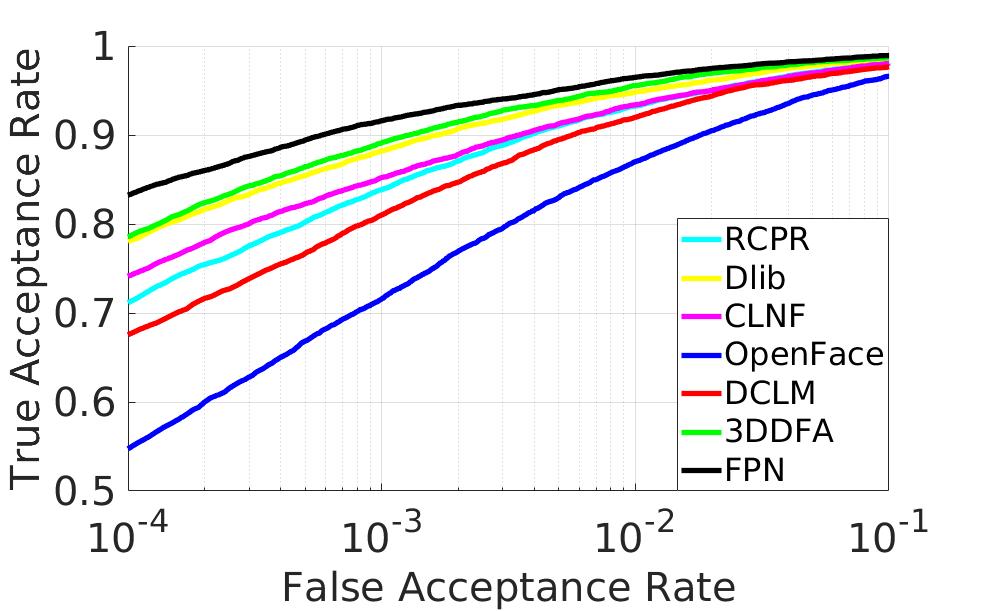}&
\includegraphics[width=.42\linewidth]{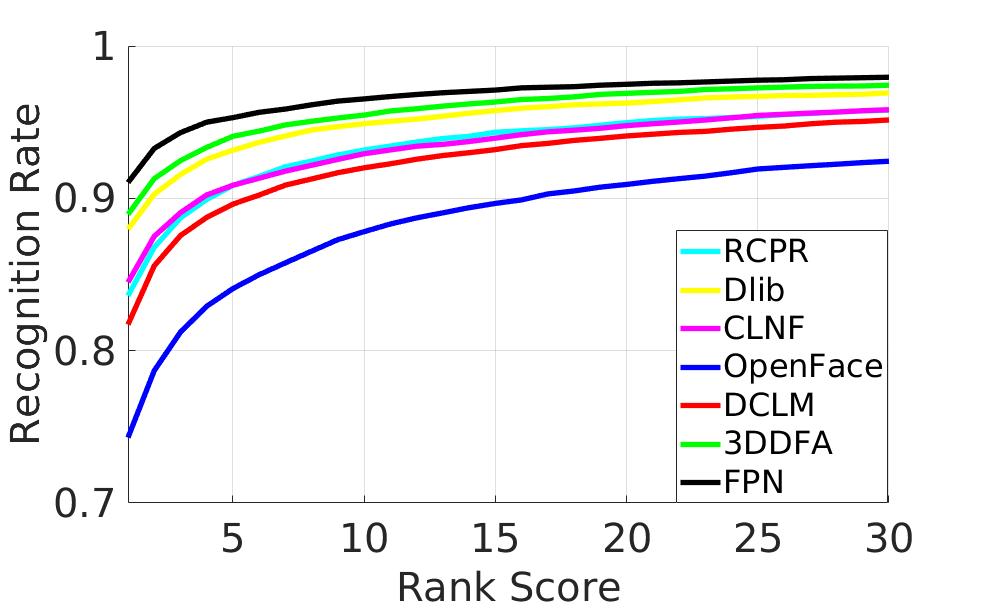}\\
 (c) ROC IJB-B & (d) CMC IJB-B\\
\end{tabular}
\caption{{\em Verification and identification results on IJB-A and IJB-B.}
ROC and CMC curves accompanying the results reported in Table~\ref{tab:IJBAB}. }\label{fig:ROC_CMC}
\vspace{-3mm}
\end{figure*}

Specifically, we use two of the most recent benchmarks for face recognition: IARPA Janus Benchmark A~\cite{Klare_2015_CVPR} and B~\cite{whitelam2017iarpa} (IJB-A and IJB-B). Importantly, these benchmarks were designed with the specific intention of elevating the difficulty of face recognition. This heightened challenge is reflected by, among other factors, an unprecedented amount of extreme out of plane rotated faces including many appearing in near-profile views~\cite{masi16dowe}. As a consequence, these two benchmarks not only push the limits of face recognition systems, but also the alignment methods used by these systems, possibly more so than the faces in standard facial landmark detection benchmarks.

\minisection{Face recognition pipeline.}
We employ a system similar to the one recently proposed by~\cite{masi2017rapid,masi16dowe}, building on their publicly available ResFace101 model and related code. We chose this system, as it explicitly aligns faces to multiple viewpoints, including rendering novel views. These steps are highly dependent on the quality of alignment and so its recognition accuracy should reflect alignment accuracy. In practice, we used their 2D (similarity transform) and 3D (new view rendering) code directly, changing how the transformations are computed: our tests compare different landmark detectors used to recover the 6DoF head pose required by their warping and rendering method, with the 6DoF regressed using our FPN. 

Their system uses a single Convolutional Neural Network (CNN), a ResNet-101 architecture~\cite{He_2016_CVPR}, trained on both real face images and synthetic, rendered views. We fine tune the ResFace101 CNN using $L2$-constrained Softmax Loss~\cite{ranjan2017l2} instead of the original softmax used by Masi et al. for their publicly released model. This fine tuning is performed using the MS-Celeb face set~\cite{liu2015faceattributes} as an example set. Aside from this change, we use the same recognition pipeline from~\cite{masi2017rapid} and we refer to that paper for details.  

\minisection{Bounding box detection.} We emphasize that an identical pipeline was used with the different alignment methods; different results vary only in the method used to estimate facial pose. The only other difference between recognition pipelines was in the facial bounding box detector.

Facial landmark detectors are sensitive to the face detector they are used with. We therefore report results obtained when running landmark detectors with the best bounding boxes we were able to determine. Specifically,  FPN was applied to the bounding boxes returned by the detector of Yang and Nevatia~\cite{yang2016multi}, following expansion of its dimensions by 25\%. Most detectors performed best when applied using the same face detector, without the 25\% increase. Finally, 3DDFA~\cite{zhu2015} was tested with the same face detector followed by the face box expansion code provided by its authors.

\begin{figure*}[t]
\centering
\subfloat[Quantitative results]{
\raisebox{1.5\height}{
\resizebox{.55\linewidth}{!}{
\begin{tabular}{lcccccc}
\toprule
\textbf{Method} & \textbf{$\le$ 5\%} & \textbf{$\le$ 10\%} & \textbf{$\le$ 20\%} & \textbf{$\ge$ 40\%} & \textbf{MER} & \multicolumn{1}{c}{\textbf{Sec./im.}} \\ \hline
RCPR~\cite{burgos2013robust} & 44.44 \% & 66.96 \% & 77.39 \% & 9.55 \% & 0.1386 &  0.19 \\ 
Dlib~\cite{king2009dlib} & 60.03 \% & 82.65 \% & 90.94 \% & 2.83 \% & 0.0795 & 0.009 \\ 
CLNF~\cite{baltrusaitis2013constrained} & 20.86 \% & 65.11 \% & 87.62 \% & 2.63 \% & 0.1106 & 0.38 \\ 
OpenFace~\cite{baltruvsaitis2016openface} & 54.39 \% & 86.74 \% & 95.42 \% & 1.27 \% & 0.0702 & 0.31 \\ 
DCLM~\cite{zadeh2016deep} & 64.91 \% & 91.91 \% & 96.00 \% & 1.17 \% & 0.0611 & 15.83 \\ 
3DDFA~\cite{zhu2015} & N/A & N/A & N/A & N/A & N/A & 0.6\\ \hline
Our FPN & 1.75 \% & 65.40 \%  & 93.86 \% & 0.97 \% & 0.1043 & 0.005 \\ 
\bottomrule
\end{tabular}
\label{fig:300w:errors}
}
}
}
\subfloat[Acumulative error curves]{
\includegraphics[width=.40\linewidth,clip,trim = 0mm 0mm 0mm 0mm]{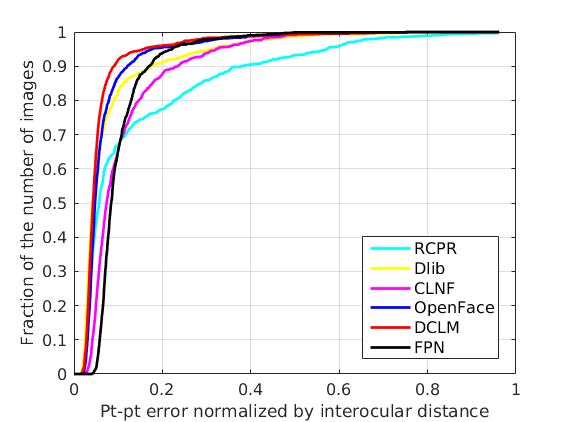}%
\label{fig:300w:qualitative}
}
\caption{{\em 68 point detection accuracies on 300W.} (a) The percent of images with 68 landmark detection errors lower than 5\%, 10\%, and 20\% inter-ocular distances, or greater than 40\%, mean error rates (MER) and runtimes. Our FPN was tested using a GPU. On the CPU, FPN runtime was 0.07 seconds. 3DDFA used the AFW collection for training. Code provided for 3DDFA~\cite{zhu2015} did not allow testing on the GPU; in their paper, they claim GPU runtime to be 0.076 seconds. As AFW was included in our 300W test set, landmark detection accuracy results for 3DDFA were excluded from this table. (b) Accumulative error curves.}
\label{fig:300w}
\vspace{-3mm}
\end{figure*}

\minisection{Face verification and identification results.}
Face verification and identification results on both IJB-A and IJB-B are provided in Table~\ref{tab:IJBAB}. We report multiple recognition metrics for both verification and identification: For verification, these measure the recall (True Acceptance Rate) at three cut-off points of the False Alarm Rate (TAR-\{1\%,0.1\%,0.01\%\}). For identification we provide recognition rates at four ranks from the CMC (Cumulative Matching Characteristic). The overall performances in terms of ROC and CMC curves are shown in Fig.~\ref{fig:ROC_CMC}. The table also provides, as reference, three state-of-the-art IJB-A results~\cite{crosswhite2017template,masi2017rapid,ranjan2017l2} and baseline results from~\cite{whitelam2017iarpa} for IJB-B (to our knowledge, we are the first to report verification and identification accuracies on IJB-B).

Faces aligned with our FPN offer higher recognition rates, even compared to the most recent, state-of-the-art facial landmark detection method of~\cite{zadeh2016deep}. In addition, our verification scores on IJB-A outperform the scores reported for the system used here as the basis for our recognition system~\cite{masi2017rapid}. These superior results are likely due to the better alignment of the faces provided by our FPN.

\begin{figure*}[t]
\centering 
\includegraphics[width=\linewidth]{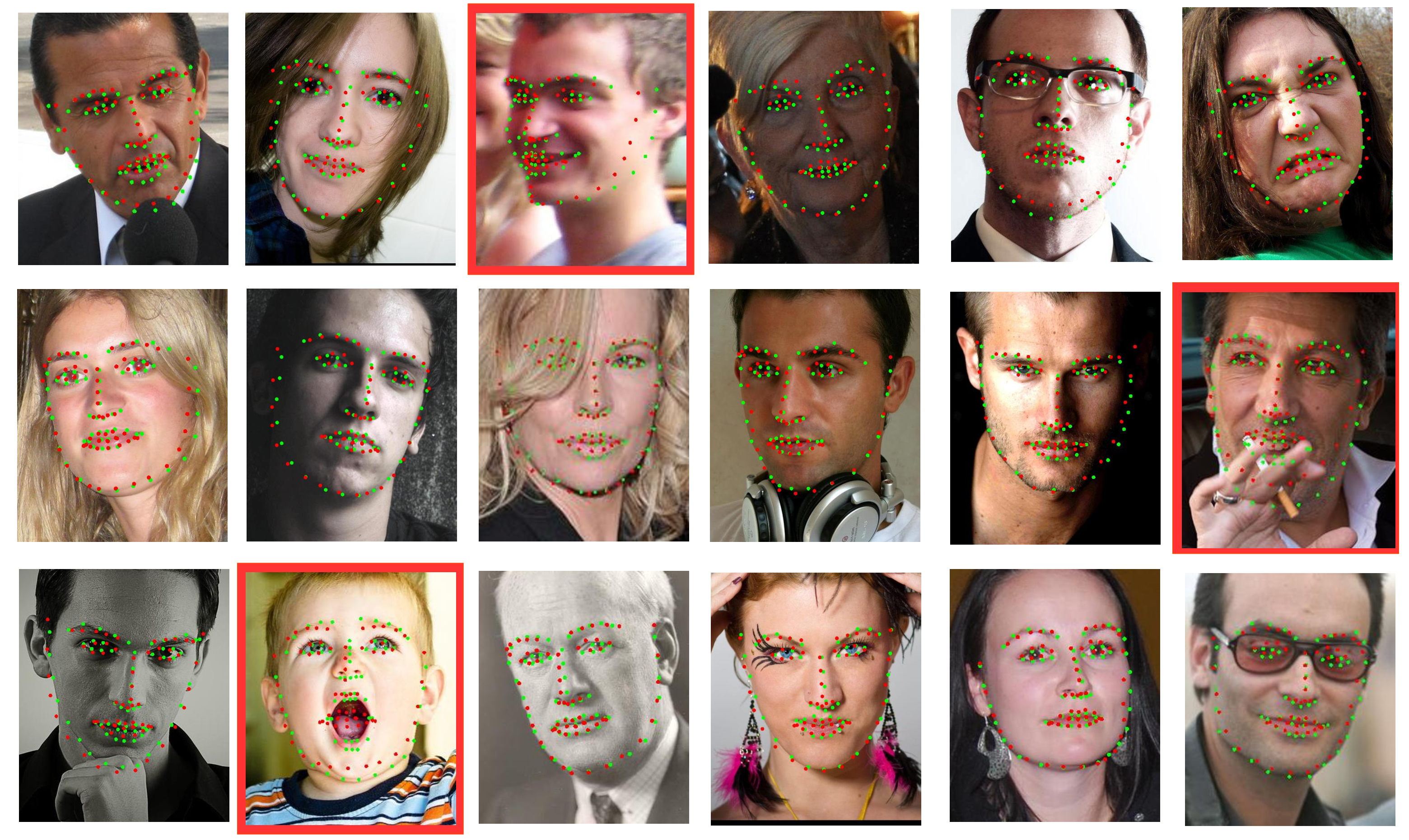}
\caption{
{\em Qualitative landmark detection examples.} Landmarks detected in 300W~\cite{sagonas2015300} images by projecting an unmodified 3D face shape, pose aligned using our FPN (red) vs. ground truth (green). The images marked by the red-margin are those which had large FPN errors ($>10\%$ inter-ocular distance). These appear perceptually reasonable, despite these errors. The mistakes in the red-framed example on the third row was clearly a result of our FPN not representing expressions.}\label{fig:FPN_vs_gt}
\vspace{-3mm}
\end{figure*}

\subsection{Landmark detection accuracy}\label{sec:300W}
\minisection{From 6DoF pose to facial landmarks.}
Given a 6DoF head pose estimate, facial landmarks can then be estimated and compared with existing landmark detection methods for their accuracy on standard benchmarks. To obtain landmark predictions, 3D reference coordinates of facial landmarks are selected off line once on the same {\em generic}, 3D face model used in~\cite{masi2017rapid}. Given a pose estimate, we convert it to a projection matrix and project these 3D landmarks down to the input image. 

Recently, a similar process was proposed for accurate landmark detection across large poses~\cite{zhu2015}. In their work, an iterative method was used to simultaneously estimate a 3D face shape, including facial expression, and project its landmarks down to the input image. Unlike them, our tests use a single generic 3D face model, unmodified. By not iterating over the face shape, our method is simpler and faster, but of course, our predicted landmarks will not reflect different 3D shapes and facial expressions. We next evaluate the effect this has on landmark detection accuracy.

\minisection{Detection accuracy on the 300W benchmark.} We evaluate performance on the 300W data set~\cite{sagonas2015300}, the most challenging benchmark of its kind~\cite{wu2015facial}, using 68 landmarks. We note that we did not use the standard training sets used with the 300W benchmark (e.g., the HELEN~\cite{le2012interactive} and LFPW~\cite{belhumeur2013localizing} training sets with their manual annotations). Instead we trained FPN with the estimated landmarks, as explained in Sec.~\ref{sec:training}. As a test set, we used the standard union consisting of the LFPW test set (224 images), the HELEN test set (330), AFW~\cite{zhu2012face} (337), and IBUG~\cite{sagonas2013300} (135). These 1026 images, collectively, form the 300W test set. Note that unlike others, we did not use AFW to train our method, allowing us to use it for testing. 

Fig.~\ref{fig:300w} (a) reports five measures of accuracy for the various methods tested: The percent of images with 68 landmark detection errors lower than 5\%, 10\%, and 20\% inter-ocular distances, and the mean error rate (MER), averaging Eq.~(\ref{eq:error}) over the images tested. Fig.~\ref{fig:300w} (b) additionally provides accumulative error curves for these methods. 

Not surprisingly, without accounting for face shapes and expressions, our predicted landmarks are not as accurate as those predicted by methods which are influenced by these factors. Some qualitative detection examples are provided in Fig.~\ref{fig:FPN_vs_gt} including a few errors larger than $10\%$. These show that mistakes can often be attributed to FPN not modeling facial expressions and shape. One way to improve this would be to use a single-view 3D face shape estimation method~\cite{hassner2013single,tran16_3dmm_cnn} to better approximate landmark positions, though we have not tested this here. 

\minisection{Detection runtime.} In one tested measure FPN far outperforms its alternatives: The last column of Fig.~\ref{fig:300w} (a) reports the mean, per-image runtime for landmark detection. {\em Our FPN is an order of magnitude faster than nearly all other face alignment methods}. Dlib~\cite{king2009dlib} was slightly slower than our FPN, but is far less accurate in the face recognition tests (Table~\ref{tab:IJBAB}). 

All methods were tested using an NVIDIA, GeForce GTX TITAN X, 12GB RAM, and an Intel(R) Xeon(R) CPU E5-2640 v3 @ 2.60GHz, 132GB RAM. The only exception was 3DDFA~\cite{zhu2015}, which required a Windows system and was tested using an Intel(R) Core(TM) i7-4820K CPU @ 3.70GHz (8 CPUs), 16GB RAM, running 8 Pro 64-bit.

\subsection{Discussion}
Landmarks predicted using FPN in Sec.~\ref{sec:300W} were less accurate than those estimated by other methods. How does that agree with the better face recognition results obtained with images aligned using FPN? As we mentioned in Sec.~\ref{sec:critique} better accuracy on a face landmark detection benchmark reflects many things which are not necessarily important when aligning faces for recognition. These include, in particular face shapes and expressions, the latter can actually cause misalignments when computing face pose and warping the face accordingly. FPN, on the other hand, ignores these factors, instead providing a 6DoF pose estimates at breakneck speeds, directly from image intensities. 

An important observation is that despite being trained with labels generated by OpenFace~\cite{baltruvsaitis2016openface}, recognition results on faces aligned with FPN are {\em better} than those aligned with OpenFace. This can be explained in a number of ways: First, FPN was trained on appearance variations introduced by augmentation, which OpenFace was not necessarily designed to handle. Second, poses estimated by FPN were less corrupted by expressions and facial shapes, making the warped images better aligned. Third, as was recently argued by others~\cite{tran16_3dmm_cnn}, CNNs are remarkably adapt at training with label noise such as any errors in the poses predicted by OpenFace for the ground truth labels. Finally, CNNs are highly capable of domain shifts to new data, such as the extremely challenging views of the faces in IJB-A and IJB-B. 

\section{Conclusions}\label{sec:conclusions}
For many practical purposes, face alignment requires only global, parametric 2D or 3D transformations. This is often the case in state-of-the-art face recognition pipelines and a wide variety of other face understanding tasks. In such circumstances, accurate facial landmark detection is superfluous and its potential for introducing errors whenever facial expressions and shapes are not explicitly considered was never fully explored. In this paper we present an alternative method for aligning faces: using a simple CNN, uniquely trained to regress 6DoF face pose, directly from image intensities. We show that by using a GPU, this leads to staggering alignment speeds. Moreover, by comparing alignment methods by considering bottom line performance of a face recognition system, rather than landmark detection accuracy, we show that this simple method outperforms state-of-the-art alignment techniques in the face recognition accuracy it provides.

\section*{Acknowledgments}
This research is based upon work supported in part by the Office of the Director of National Intelligence (ODNI), Intelligence Advanced Research Projects Activity (IARPA), via IARPA 2014-14071600011. The views and conclusions contained herein are those of the authors and should not be interpreted as necessarily representing the official policies or endorsements, either expressed or implied, of ODNI, IARPA, or the U.S. Government.  The U.S. Government is authorized to reproduce and distribute reprints for Governmental purpose notwithstanding any copyright annotation thereon.

{\small
\bibliographystyle{ieee}
%\bibliography{faceposenet}

}

\end{document}